\documentclass[runningheads]{llncs}
\usepackage[framemethod=tikz]{mdframed}

\usepackage[T1]{fontenc}
\usepackage{appendix}
\usepackage{xurl}
\usepackage{url}
\usepackage{float}
\usepackage{soul}
\usepackage{booktabs}
\usepackage{multirow}
\usepackage{makecell}
\usepackage{xcolor}
\usepackage{colortbl}
\usepackage{threeparttable}
\usepackage{todonotes}
\usepackage{graphicx}
\usepackage{wrapfig}
\usepackage{amssymb,amsfonts}
\usepackage{devanagari}
\usepackage{blindtext}
\usepackage{enumitem}
\usepackage{array}
\usepackage[normalem]{ulem}
\usepackage[dvipsnames]{xcolor}
\definecolor{cvprblue}{rgb}{0.21,0.49,0.74}

\usepackage{xspace}
\makeatletter
\DeclareRobustCommand\onedot{\futurelet\@let@token\@onedot}
\def\@onedot{\ifx\@let@token.\else.\null\fi\xspace}

\def\eg{\emph{e.g}\onedot}

\def\etal{\emph{et al}\onedot}
\makeatother

\let\oldcolorbox\colorbox

\renewcommand{\colorbox}[2]{%
  \begingroup
  \setlength{\fboxsep}{0pt}
  \oldcolorbox{#1}{\strut #2}%
  \endgroup
}
\newcommand{\ENHuman}{\textit{EN-Human}\xspace}
\newcommand{\HIHuman}{\textit{HI-Human}\xspace}
\newcommand{\HumanADToHindi}{\textit{Human AD-to-Hindi}\xspace}
\newcommand{\PredADToHindi}{\textit{Pred AD-to-Hindi}\xspace}
\newcommand{\PredDenseToHindi}{\textit{Pred Dense-to-Hindi}\xspace}
\definecolor{adgreen}{HTML}{CCEBC5}
\definecolor{ddred}{HTML}{F6C6AD}
\definecolor{adcyanlime}{HTML}{C0F2D9}
\definecolor{sameevent}{HTML}{C2F1C8}
\definecolor{partialevent}{HTML}{FFFFCC}
\definecolor{differentevent}{HTML}{FBB4AE}
\definecolor{translationchoice}{HTML}{83CBEB}
\definecolor{isolated}{HTML}{D9D9D9}

\newcommand{\mypara}[1]{\vspace{2mm}\noindent\textbf{#1}}
\newcommand{\dataset}{Andha-Dhun\xspace}

\usepackage[breaklinks,colorlinks,citecolor=cvprblue]{hyperref}

\usepackage[capitalize]{cleveref}
\crefname{figure}{Fig.}{Figs.}
\Crefname{figure}{Figure}{Figures}
\crefname{section}{Sec.}{Secs.}
\Crefname{section}{Section}{Sections}
\crefname{table}{Tab.}{Tabs.}
\Crefname{table}{Table}{Tables}
\begin{document}

\title{\dataset:\\A First Look at Audio Descriptions in Hindi}

\author{
Ritabrata Chakraborty*\inst{1,2} \and
Divy Kala*\inst{1} \and
Nisheeth Bhooshan Gupta\inst{3} \and \\
Ganji Sreeram\inst{3} \and
Pailla Balakrishna Reddy\inst{3} \and
Makarand Tapaswi\inst{1}}
\authorrunning{R. Chakraborty, D. Kala et al.}
\institute{
$^1$CVIT, IIIT Hyderabad,
$^2$Manipal University Jaipur,
$^3$Jio Platforms Ltd. \\
\url{https://github.com/katha-ai/AndhaDhun-HindiAD} \\
{\footnotesize $^*$ denotes equal contribution}
}

\maketitle
\begin{abstract}
Audio Descriptions (ADs) narrate visual content for Blind and Low Vision (BLV) audiences during gaps in audiovisual media.
There is growing momentum around ADs in movies and TV shows, and with mandates from India's Central Board of Film Certification (CBFC), there is a need to expand ADs beyond English.
Yet, there is no work that generates ADs for any Indian language. 
To address this gap, we present the first systematic study of ADs in Hindi, contributing to aspects such as data, generation, and evaluation.
We introduce \dataset, the first dataset of human-authored Hindi ADs collected from 8 full-length movies.
We explore two approaches for generating ADs in Hindi:
(i)~directly from English dense video descriptions, and
(ii)~translating English ADs into Hindi.
We evaluate these approaches using perplexity and LLM-as-a-judge metrics to assess fluency and quality respectively.
We also analyze movies that have both English and Hindi human-authored ADs and find that naïve translation introduces artifacts and narrows diversity compared to original Hindi ADs.
Direct machine translation fails to adapt cultural references, while human-translated ADs do better but still fall short.
Our findings emphasize that the purpose of Hindi ADs is accessibility for Indian BLV audiences, and that this requires adapting content for the audience more than strict fidelity to the source.

\end{abstract}

\setcounter{footnote}{0}

\section{Introduction}
\label{sec:introduction}

\begin{figure}[t]
\centering
\includegraphics[width=0.99\linewidth]{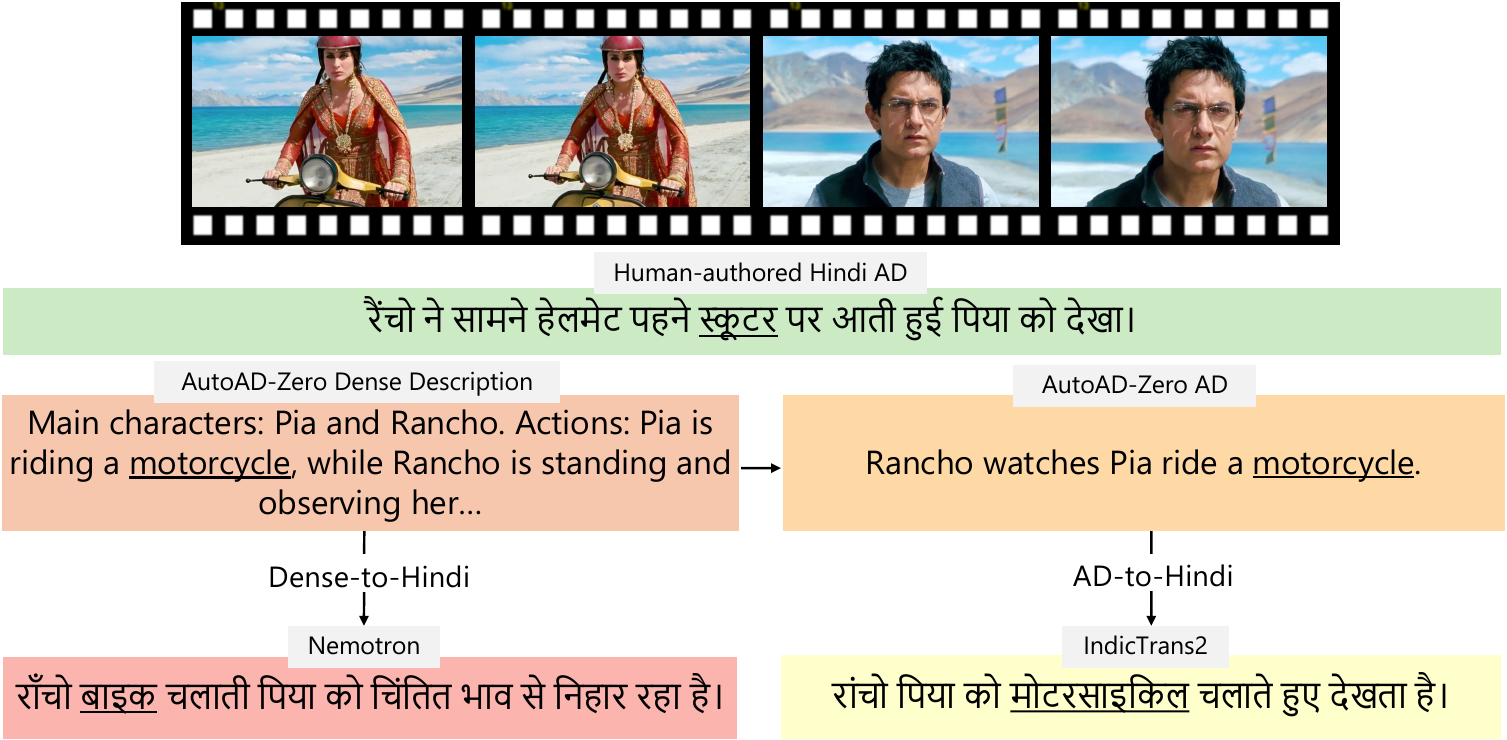}
\vspace{-2mm}
\caption{
\textbf{Example of a Hindi audio description}
showing a clip from \textit{3 Idiots} with the human-authored Hindi AD.
AutoAD-Zero, an automatic AD generation framework, provides a
dense description
of the scene in English, which is then summarized to an
English AD
by an LLM.
We show examples of two methods to get Hindi ADs:
(bottom-left) \underline{Dense-to-Hindi}, obtained by using a Hindi capable LLM directly on the English dense descriptions; and (bottom right) \underline{AD-to-Hindi}, where AutoAD-Zero's~\cite{autoadzero} English AD is translated to Hindi using multiple translation models.
The example shows an instance where a culture-specific item such as \textit{scooter} is wrongly perceived as a motorcycle, resulting in downstream errors in Hindi generation.
}
\vspace{-4mm}
\label{fig:teaser}
\end{figure}

Audio Descriptions (ADs) are narrations inserted into silent gaps (\eg~between dialog) to help Blind and Low Vision (BLV) audiences understand visual content in movies
and TV shows.
An example AD is shown in \cref{fig:teaser}.
A sighted viewer perceives the different modes (visuals, music, speech, and sound effects) of a movie jointly to create an experience of a \textit{coherent whole}.
It is the job of an expert audio describer, to identify and isolate that information which is not accessible to the BLV audience, but is essential in creating that \textit{coherent whole}~\cite{braun2008audiodescription}.
Consequently, creating ADs requires expert knowledge and is a time-consuming and expensive process, taking up to 35-40 working hours for a 90-minute movie~\cite{adsurveyllmvlm}.

Growing support for ADs has led to international regulatory efforts~\cite{europeanADparliament}, and this momentum has led to an explosion of interest in automatic AD generation for English~\cite{distinctad,autoad3,narrad,shotbyshot,autoadzero}
and European languages~\cite{mazur2012towards,orero2007sampling,machinetranslationADpostediting,adsurveyllmvlm}.
Similarly in India, the Central Board of Film Certification (CBFC) announced that all movies releasing in India will be required to have ADs~\cite{CBFC}. 
However, there is no prior work on ADs for Indian languages.
In fact, the ADP ACB Title Directory%
\footnote{ADP Title Directory: \url{https://adp.acb.org/adp-search} is an initiative of the American Council of the Blind, providing a comprehensive directory of expert written ADs for content in multiple languages.} 
lists only around 650 titles of ADs in Hindi, compared to over 13.5K for English.

A natural solution to this problem is to translate existing ADs in English (or other languages) into Hindi.
However, prior studies on AD translation across European languages show that machine-translated ADs often require human correction and cannot be assumed to substitute for the target-language ADs directly~\cite{machinetranslationADpostediting,englishdutchAD,dialnet,lfrenchgermanAD}. 
They also highlight specific issues with machine-translated ADs, including post-editing burden~\cite{machinetranslationADpostediting}, machine-translation errors, timing mismatches~\cite{englishdutchAD}, and the lack of visually grounded, time-aligned AD data~\cite{fischer-etal-2025-swissadt}.

Skopos theory~\cite{vermeer1996skopos} (from the German word for ``purpose'') offers insights to understand why.
It states that a translation should be guided by its \textit{skopos}; for Hindi AD, this means creating descriptions that fulfill the purpose of \textit{helping Indian BLV audiences understand the movie and its visuals}, rather than strictly following the original wording.
For instance, consider the English AD: \textit{``He scored a touchdown and celebrated with a Gatorade shower.''}
A naïve translation to Hindi would be \textit{``Usne \underline{touchdown} score kiya aur \underline{Gatorade shower} ke saath celebrate kiya.''} 
However, this preserves culture-specific terms that may mean little to a Hindi-speaking BLV audience.
In contrast, a skopos-aware translation might read \textit{``Usne \underline{winning score} banaya aur teammates ne \underline{uspe drinks daal kar} celebrate kiya''}, fulfilling the purpose of the AD for the target audience.

We study the problem of Hindi ADs on three fronts:
data, generation, and evaluation, all of which have not been explored before.
We examine the effectiveness of translation models for converting human-authored English ADs into Hindi, and note the introduction of translation artifacts and a reduction in diversity.
We show that the Hindi AD landscape is still at an early stage for both generation and translation systems.
Furthermore, they often fail to adapt content for Indian BLV audiences, particularly when handling cultural references.
Our contributions are as follows:
\begin{enumerate}
\item \textbf{Data.}
We present \dataset\footnote{\textit{Andha} means `blind' and \textit{dhun} means `melody' (or `sound') in Hindi.}, a dataset of 5,870 AD sentences across 8 full-length movies. This is the first corpus for studying ADs in Hindi.
\item \textbf{Generation.} We investigate the automatic generation of ADs in a zero-shot setup~\cite{autoadzero} and explore two translation-based approaches:
(i)~directly generate ADs from English dense descriptions using Hindi-capable models, and (ii)~translate English ADs into Hindi.
\item \textbf{Evaluation.}
We evaluate the quality of Hindi ADs using intrinsic (perplexity) and extrinsic (LLM-as-a-judge) metrics.
We also study artifacts of Culture-Specific Items (CSIs) in machine- and human-translated ADs.
\end{enumerate}

\section{Related Work}
\label{sec:related_work}

\mypara{Accessibility and translation.}
Early research~\cite{visualinformationAD,audetel} introduced ADs as an accessibility problem, emphasizing comprehension and usability for BLV users to follow audiovisual content.
With rising adoption of ADs in movies after the European Parliament accessibility directives (2007)~\cite{EUDirective2007_65}, movie-aware and translation-aware work gained interest~\cite{machinetranslationADpostediting,adsurveyllmvlm}.
Handling culture-bound~\cite{cultureboundelements,culturalreferences} elements was a recurring problem for foreign movie ADs.
To this end, many works~\cite{skoposAD,skoposADforeignfilms} focused on using Skopos theory~\cite{skoposintroduction}, to emphasize the \textit{purpose} of ADs, and keeping in mind the target demographic.
The Skopos framework connected low resource machine translation research~\cite{machinetranslationsurvey} to AD translations.
Across multiple settings, Vercauternen~\etal~\cite{englishdutchAD,machinetranslationADpostediting} showed that  mistranslations remain a major limitation in simply using translated ADs, hence human post-editing is needed.

\mypara{Automatic AD generation.}
With the rise of Multimodal LLMs (MLLMs), automatic AD generation has grown in popularity~\cite{autoad3,maddataset,autoadzero,adsurveyllmvlm}.
Many approaches are training-based~\cite{distinctad,autoad,autoad2,autoad3,interleavedAD,contextualAD}, while a growing body of work explores zero-shot methods~\cite{llmad,autoadzero,mmad,mmnarrator,coherentAD,shotbyshot,narrad}.
Typically, zero-shot methods first use an MLLM to produce dense visual descriptions, which are then transformed into ADs by an LLM~\cite{shotbyshot,autoadzero}.
In recent work, Shot-by-Shot~\cite{shotbyshot} uses shot scale and film-grammar to improve AD generation, while CoherentAD~\cite{coherentAD} uses a multi-stage approach to ensure ADs are coherent.
Across multiple training-based and zero-shot methods, generated ADs consistently show a significant gap from achieving human-authored AD quality~\cite{braunevaluation,adqa}.
All this work is catered primarily towards English language AD generation.

\mypara{Relevance of our work.}
We present a first work on AD generation for an Indian language (Hindi).
Building on AutoAD-Zero~\cite{autoadzero}, we evaluate translation-based AD generation.
We also analyze shortcomings of using translations, both for automatically generated and human-authored English ADs.

\section{The \dataset Dataset}
\label{sec:dataset}

\begin{figure}[t]
\centering
\includegraphics[width=0.99\linewidth]{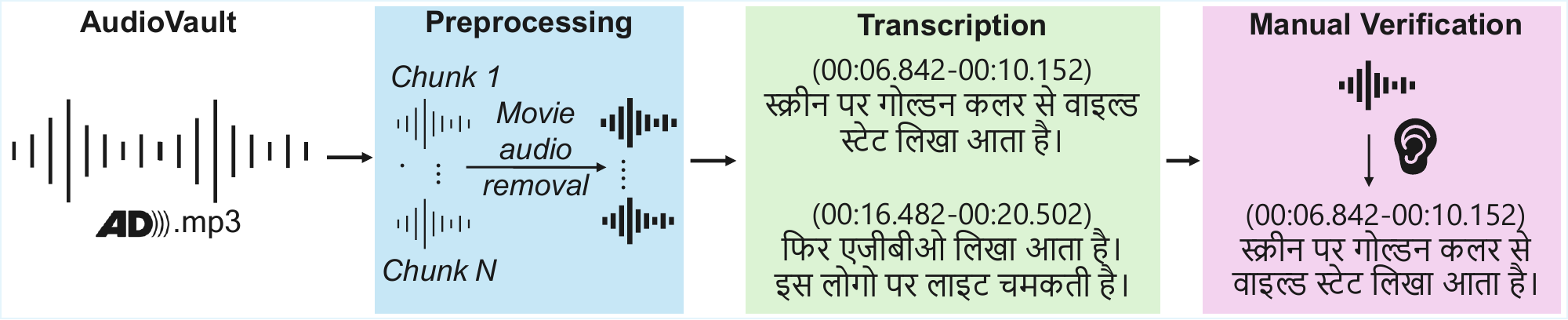}
\vspace{-2mm}
\caption{\textbf{\dataset dataset collection pipeline.}
Given movie ADs in audio format, we perform some preprocessing (chunking and movie audio removal).
The cleaned audio chunks are transcribed using Gemini-2.5-Flash to obtain timestamped ADs.
A final manual verification ensures high quality transcription and temporal alignment.}
\label{fig:dataset_pipeline}
\vspace{-4mm}
\end{figure}

To create the dataset, we source the AD audio of a movie from AudioVault\footnote{A non-profit repository for ADs, available at \url{https://audiovault.net}.}, where each file contains the original movie soundtrack and Hindi AD narrations.
We split each movie into roughly 30-minute chunks and process them with Resemble Enhance\footnote{A tool that improves quality of speech by performing denoising and enhancement, available at \url{https://huggingface.co/ResembleAI/resemble-enhance}} to suppress non-AD sound sources, producing narrator-focused audio for transcription.
Each chunk is transcribed with Gemini-2.5-Flash, which returns Hindi transcripts with start and end timestamps and labels each spoken segment as either actor (dialog) or narrator (AD).
All outputs are manually verified and corrected to fix transcription errors and inaccurate boundaries. 
\cref{fig:dataset_pipeline} summarizes this pipeline and the transcription prompt is shown in \cref{app:dataset}.

To align the transcribed ADs with visuals, we source the corresponding movie videos from DVDs or public platforms.
We manually synchronize the cleaned Hindi AD audio with the visual stream.
Thus, the \dataset{} dataset consists of transcribed ADs synchronized with movie visuals.
Since existing AD generation methods require a character bank~\cite{autoadzero,autoad2,autoad3,shotbyshot} for identifying characters in the movie, we also create character banks for all movies using IMDb\footnote{IMDb (Internet Movie Database), available at \url{https://www.imdb.com/}\label{imdbfootnote}} cast profiles. 
\cref{tab:hindi_ad_per_movie_stats_compact} provides some statistics of the dataset.

\begin{table}[t]
\caption{\textbf{An overview of the \dataset dataset} consisting of 4 Hindi movies with Hindi ADs and
4 English movies with both Hindi \textit{and} English ADs.
The latter facilitate studying effects of translation.
AD duration is the narration time in seconds.}
\label{tab:hindi_ad_per_movie_stats_compact}
\vspace{-2mm}
\centering
\small
\tabcolsep=0.16cm
\begin{tabular}{c l cc c c}
\toprule
\multirow{2}{*}{Source} &
\multirow{2}{*}{Movie} & 
\multicolumn{2}{c}{Duration (s)} &
\multirow{2}{*}{\#ADs} & Average \\
& & Movie & AD & & words/AD \\
\midrule
\multirow{4}{*}{\rotatebox[origin=c]{90}{Hindi}} 
& 3 Idiots    & 10,234 & 2,589 & 761 & 18.8 \\
& Airlift     & 7,385 & 2,041 & 686 & 16.3 \\
& Gehraaiyan  & 8,880 & 1,629 & 606 & 15.1 \\
& Lage Raho Munna Bhai & 8,707 & 1,428 & 516 & 14.6 \\
\midrule
\multirow{4}{*}{\rotatebox[origin=c]{90}{English}} 
& Slumberland    & 7,237 & 2,362  & 895 & 16.9 \\
& Murder Mystery & 5,912 & 1,197  & 499 & 15.0 \\
& Heart of Stone & 7,525 & 3,167  & 997 & 17.7 \\
& Extraction 2   & 7,437 & 2,489  & 910 & 13.2 \\
\midrule
& Total (8 movies) & 63,317 & 16,902 & 5,870 & 16.1 \\
\bottomrule
\end{tabular}
\vspace{-3mm}
\end{table}

\section{Automatic Generation of Hindi ADs}
\label{sec:generation}

As current English AD generation methods operate  on few-second trimmed clips~\cite{maddataset,autoad3}, we convert our dataset into video-AD pairs containing a short video clip and associated Hindi AD.
We use these pairs for AD generation.

\subsection{Experimental Setup and Metrics}
\label{subsec: experimental_setup}

\mypara{Zero-shot AD generation.}
We adopt AutoAD-Zero~\cite{autoadzero}, a zero-shot AD generation framework that uses an off-the-shelf MLLM to generate a dense description of a few-second video clip, which is then distilled into an AD by an LLM.
To improve character grounding, the framework augments the visual frames with cast information predicted based on a character bank from IMDb cast profiles\footref{imdbfootnote}.
Qwen-2-VL-7B-Instruct~\cite{qwen2vltechnicalreport} is the MLLM and Llama-3-8B is the LLM~\cite{grattafiori2024llama3herdmodels}.

\begin{figure}[t]
\centering
\includegraphics[width=0.99\linewidth]{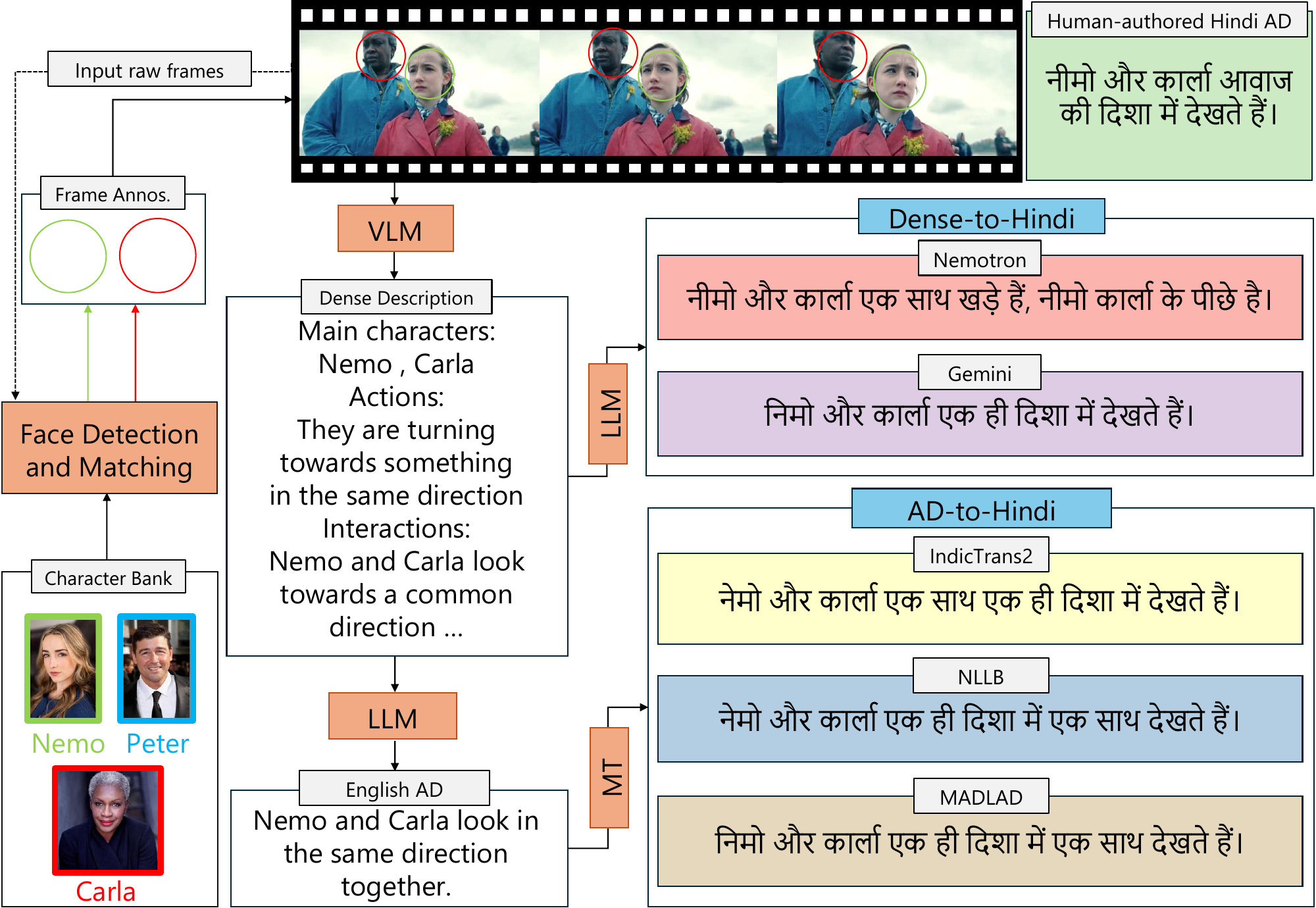}
\vspace{-2mm}
\caption{\textbf{Zero-shot Hindi AD generation pipeline}.
(From bottom-left)
(i)~A character bank of the movie cast is used to perform face recognition and create in-frame annotations for the input video.
(ii)~The annotated frames are passed to an MLLM to generate a dense description, which is then passed to an LLM to get a summarized English AD.
This is the AutoAD-Zero pipeline~\cite{autoadzero}.
(iii)~To obtain Hindi ADs, we propose two approaches:
(bottom right) \textit{AD-to-Hindi} uses a machine translation model to translate the English AD to Hindi; and
(middle right) \textit{Dense-to-Hindi} uses a Hindi-capable LLM to obtain Hindi ADs directly from the English dense descriptions.
For reference the human-authored Hindi AD is also shown (top-right).}
\label{fig:generation}
\vspace{-4mm}
\end{figure}

\mypara{AD generation approaches.} For generating ADs in Hindi, we explore two approaches (see \cref{fig:generation}):
(i)~\textbf{Pred AD-to-Hindi}, where we first generate English ADs using the AutoAD-Zero method and then translate them into Hindi using machine translation models such as IndicTrans2~\cite{indictrans2}, No Language Left Behind (NLLB-600M)~\cite{nllb}, or MADLAD~\cite{madlad}, and a commercial LLM-based translation using Gemini-3.1-Pro.
(ii)~\textbf{Pred Dense-to-Hindi} distills the MLLM generated dense-descriptions into Hindi ADs directly by using a Hindi-capable LLM.
We use Nemotron-4-Mini-Hindi-4B-Instruct~\cite{nemotronhindi} and Gemini-3.1-Pro~\cite{geminiteam2025geminifamilyhighlycapable} as the Hindi-capable LLMs.
For \PredDenseToHindi with Gemini, we provide 5 previous ADs and/or dialogs to establish narrative context for the LLM.
The prompt for Dense-to-Hindi is shown in \cref{app:machine_generated_ad}.

\mypara{Metrics.}
We evaluate the quality of generated ADs from two complementary perspectives: extrinsic and intrinsic.
As an extrinsic metric, we use \textit{LLM-AD-eval}~\cite{autoad3}, an LLM-as-a-judge framework~\cite{li2024llmasajudge} that prompts a language model to evaluate the similarity between a predicted and ground-truth AD, assigning a score from 0 (worst match) to 5 (best match).
The LLM-AD-Eval prompt is shown in \cref{app:machine_generated_ad}.
As an intrinsic metric, we use \textit{perplexity}~\cite{perplexity1977}, which measures how predictable the ADs are for a language model. 
Perplexity does not require ground-truth and acts as a measure of linguistic diversity~\cite{garbacea-etal-2019-judge,hashimoto-etal-2019-unifying}.

The same metrics are also used in assessing human-authored ADs in \cref{sec:analysis}.
Given the recent findings by Kala~\etal~\cite{adqa}, we do not use classical captioning metrics such as CIDEr.
Additionally, as Hindi allows flexible word order to convey the same meaning, two sentences can be semantically identical despite having low n-gram overlap~\cite{bleuhindi,en2hi_mt}.
Hence, surface-level metrics like CIDEr and BLEU may be unreliable for evaluating Hindi generations.

\begin{table}[t]
\centering
\tabcolsep=0.15cm
\caption{\textbf{LLM-AD-Eval scores for predicted ADs}.
Under both approaches (\PredADToHindi and \PredDenseToHindi), we evaluate quality with two different models in LLM-as-a-judge setup: Gemini-3.1-Pro \cite{geminiteam2025geminifamilyhighlycapable} and LLaMA-3-8B-Instruct \cite{grattafiori2024llama3herdmodels}. \textbf{Best results} and \underline{second-best results} per translation mode (along each row) are highlighted.
}
\label{tab:machine_ad_quality_merged}
\vspace{-2mm}
\begin{tabular}{l | cccc | cc}
\toprule
& \multicolumn{4}{c|}{AD-to-Hindi} & \multicolumn{2}{c}{Dense-to-Hindi} \\
Judge & IndicTrans2 & NLLB & MADLAD & Gemini & Nemotron & Gemini w/ ctxt \\
\midrule
Llama  & \underline{3.07} & 3.03 & 2.91 & 2.99 & \textbf{3.38} & 3.00 \\
Gemini & 1.00 & 0.99 & 0.84 & 0.94 & \textbf{1.12} & \underline{1.09} \\
\bottomrule
\end{tabular}
\vspace{-2mm}
\end{table}

\subsection{Results}

\begin{figure}[t]
\centering
\includegraphics[width=0.6\linewidth]{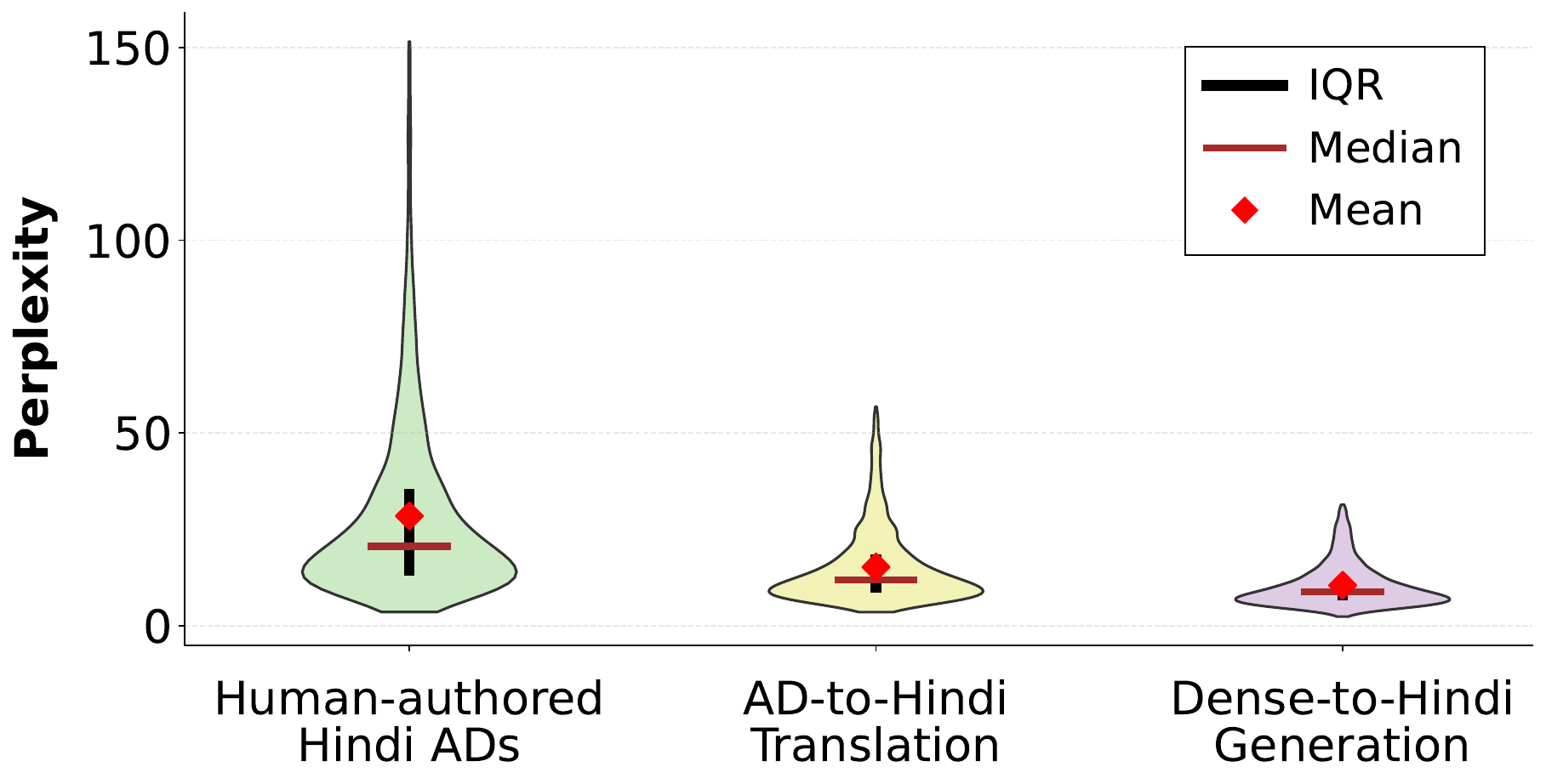}
\vspace{-2mm}
\caption{\textbf{Perplexity of Hindi ADs.}
The violin plot shows values up to 95th percentile.
AD-to-Hindi translation uses IndicTrans2 to translate ADs, and
Dense-to-Hindi uses Nemotron to generate ADs.
The base model is Llama-3.1-8B.}
\vspace{-5mm}
\label{fig:machine_generated_perplexity}
\end{figure}

\mypara{Perplexity.}
We compute perplexity scores using Llama-3.1-8B. 
Due to a long-tail of extremely rare tokens/words, we remove outliers beyond the 95th percentile.
In \cref{fig:machine_generated_perplexity}, we observe that perplexity is highest for human-authored Hindi ADs, followed by AD-to-Hindi translation (IndicTrans2).
Dense-to-Hindi generations by Nemotron display the lowest perplexity.
This suggests that human-authored ADs have most linguistic diversity while capturing the rich visual details, while both AD-to-Hindi and Dense-to-Hindi lack this quality.

\mypara{LLM-AD-Eval.}
\cref{tab:machine_ad_quality_merged} reports LLM-AD-Eval scores for the AD generation choices described in \cref{subsec: experimental_setup}.
Among the AD-to-Hindi translation models, both judges consistently rank IndicTrans2 highest and MADLAD lowest, with NLLB and Gemini falling in between.
For the Dense-to-Hindi approach, the smaller Hindi-focused Nemo-tron achieves the best scores under both judges, outperforming the larger multilingual Gemini even with AD/dialog context.

The two judges differ substantially in their absolute scoring scales: the Gemini judge assigns scores in a narrow range around 0.84--1.12, whereas the Llama judge produces scores between 2.91 and 3.38. 
While absolute scores are different, likely due to varying harshness exhibited by the models,
they consistently identify Nemotron as the strongest and MADLAD as the weakest methods.

Interestingly, the Dense-to-Hindi Nemotron surpasses all AD-to-Hindi translation methods, suggesting that directly generating Hindi AD from dense captions can be more effective than translating predicted English ADs. 
Nevertheless, even the highest scores remain well below the upper end of the 0--5 scale, confirming that current automatic Hindi AD generation methods are still far from human-level (results for translated human-authored English ADs in \cref{subsec:trans_enhuman}).

\section{Human-Authored English to Hindi ADs}
\label{sec:analysis}

\begin{figure}[t]
\centering
\includegraphics[width=0.9\textwidth]{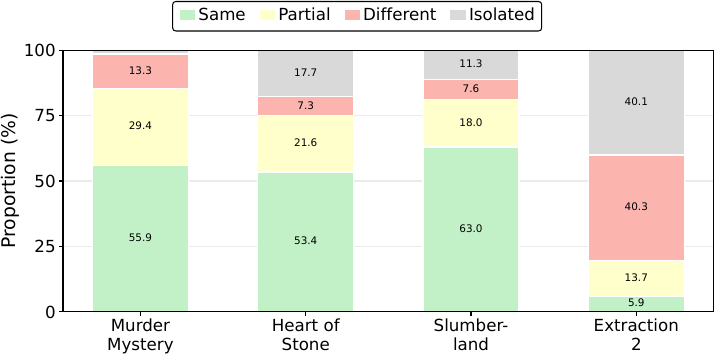}
\vspace{-2mm}
\caption{\textbf{Human-authored Hindi ADs closely match human-authored English ADs, machine translated to Hindi.}
We form pairs by temporally aligning human-authored Hindi ADs (\HIHuman) with Hindi translations of human-authored English AD sentences (\HumanADToHindi).
Each bar shows the fraction of times these paired ADs describe the \colorbox{sameevent}{same details}, \colorbox{partialevent}{partially overlapping details}, or \colorbox{differentevent}{entirely different details}.
\colorbox{isolated}{Isolated} or non-aligned cases are also displayed.
For the first three movies, we see a strong correspondence with most pairs describing the same or partially overlapping details.
On the other hand, \textit{Extraction~2} has many isolated ADs, and even the aligned ones describe different details.
}
\label{fig:dualtrack_stacked}
\vspace{-2mm}
\end{figure}

\mypara{Dual AD tracks.} 
The four English movies in \dataset{} are chosen to have human-authored ADs in both languages (\ENHuman, \HIHuman).
We transcribe the English ADs automatically%
\footnote{We transcribe the \ENHuman AD track using WhisperX~\cite{bain2022whisperx} and classify each sentence as \textit{AD} or \textit{dialog} using Gemini-3.1-Pro, following~\cite{adqa}.},
and observe that the English and Hindi ADs are closely aligned, hinting that the Hindi ADs are likely expert human translations.
Next, we create \HumanADToHindi, a set of Hindi ADs obtained by translating English ADs using automatic methods from \cref{subsec: experimental_setup}.

To understand correspondences, we manually align \HIHuman AD sentences with \ENHuman based on temporal overlap.
We first match entries by their start times across both tracks.
When durations do not align, this is often due to over-segmentation during transcription.
We merge adjacent entries until the combined duration matches its counterpart and treat the merged span as a single AD unit.
This results in 2,350 Hindi and 2,358 English ADs, of which, 2,061 time-aligned pairs are used in further analysis.

\mypara{Amount of overlapping detail in AD tracks.}
Given two paired AD sentences from the dual-track subset, we explore how much visual detail they share.
We instruct Gemini-3.1-Pro to classify each \HIHuman and \HumanADToHindi pair into three categories: 
(i)~\colorbox{sameevent}{same details}: same primary action/event even if one sentence is more detailed or uses synonyms, 
(ii)~\colorbox{partialevent}{partially overlapping details}: tracks share some visual content, but one also describes a clearly distinct additional action absent from the other, or
(iii)~\colorbox{differentevent}{different details}: completely unrelated with no meaningful overlap in action, object, or character.
The LLM also provides a one-sentence justification for each classification.
We also perform manual validation for 50 pairs per movie to confirm that the assignments are reliable.
The detailed prompt is provided in \cref{app:human_authored_ad_generation}.

\begin{figure}[t]
\centering
\includegraphics[width=0.99\linewidth]{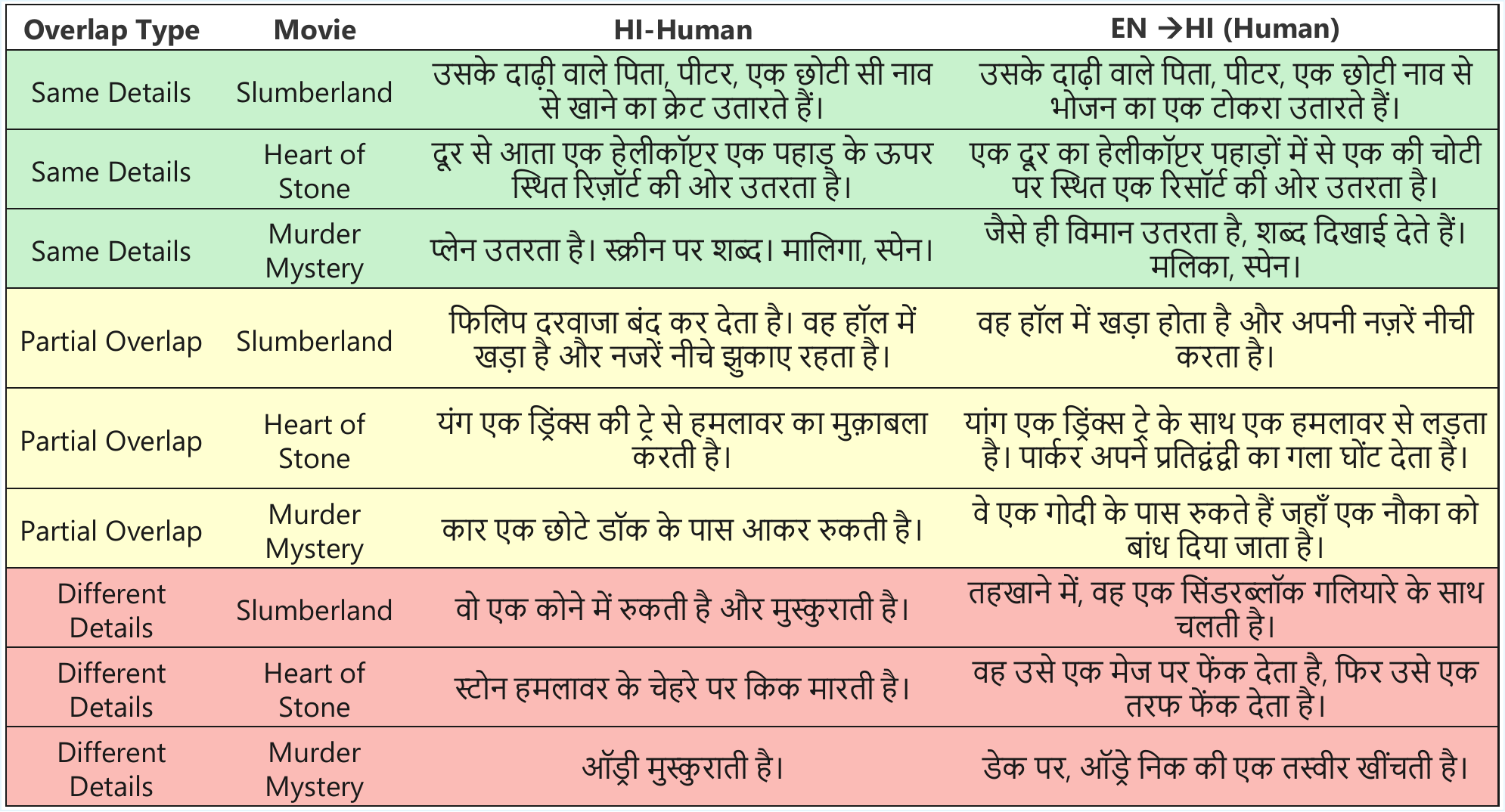}
\vspace{-2mm}
\caption{\textbf{Qualitative examples of semantic overlap in dual-track subset.}
We show aligned AD pair examples with
\colorbox{sameevent}{same details}, 
\colorbox{partialevent}{partial overlap}, and 
\colorbox{differentevent}{different details}. 
\HIHuman refers to human-authored Hindi ADs.
\textit{EN$\rightarrow$HI (Human)} refers to machine translated Hindi ADs from human-authored English ADs (\HumanADToHindi).}
\label{fig:semantic_overlap_examples}
\vspace{-3mm}
\end{figure}

\cref{fig:dualtrack_stacked} shows the breakdown for each movie and \cref{fig:semantic_overlap_examples} shows some qualitative examples.
Across 3 movies, same and partial correspondences dominate, confirming that \HIHuman ADs are largely human-edited translations of the corresponding \ENHuman ADs.
This makes them a useful reference to analyze how meaning is preserved and adapted across languages in ADs.
\textit{Extraction~2} is an outlier, since its AD tracks describe different salient details (often for fast-paced action scenes) and is hence excluded from subsequent experiments.

\subsection{Culture-Specific Items in ADs} 
\label{subsec:csi_main}

\mypara{Experimental setup.}
To ensure that the cultural references in translated ADs are genuinely accessible to an Indian BLV audience, we designed an evaluation pipeline grounded in \textit{Skopos} theory.
Here, the goal of the translation is to serve its intended purpose---accessibility for an Indian BLV audience.
This is, in part, achieved by prioritizing cultural aspects over strict source-text fidelity. 

We conduct this experiment in two phases.
In the first phase, we identify Culture-Specific Items~\cite{csis} (\eg~``quarterback'', ``ivy league colleges'', ``kitchen island'') in the \ENHuman side of the 4 movie dual track subset, which includes \textit{Extraction 2}.
Since the target audience might not understand Culture-Specific Items (CSIs) from the source culture, it is imperative to resolve them while translating.
We use Gemini-3.1-Pro\footnote{Gemini-3.1-Pro gave better results than 2.5-Pro for identifying cultural nuances.}
to flag CSIs in 2,358 unique \ENHuman ADs using the prompt given in \cref{fig:csi_detection_prompt}.
We manually verify the flagged samples to obtain 141 ADs (6.0\%) with CSIs.

In the second phase, 
we translate the \ENHuman ADs into Hindi to produce \HumanADToHindi ADs and manually evaluate whether these translations resolve the identified CSIs.
We use 120 of 141 ADs aligned with \HIHuman ADs from the dual track for further analysis.
We categorize CSI resolution strategies based on Pederson's taxonomy~\cite{pedersen2011_csi_taxonomy}:
(1)~\textit{retention}: keeping the original term, \eg~\textit{gurney};
(2)~\textit{specification}: adding clarification, \eg~\textit{pahiyen wala stretcher};
(3)~\textit{direct translation}: literal rendering, \eg~\textit{White House} as \textit{Safed Ghar}; 
(4)~\textit{generalization}: using a broader term, \eg~\textit{Five Guys} as a fast food joint;
(5)~\textit{substitution}: replacing with a cultural equivalent, \eg~\textit{as wise as an owl} with \textit{Chanakya jaisa samajhdar};
and
(6)~\textit{omission}: removing the reference altogether.
We also manually analyze the corresponding \HIHuman ADs to assess CSI resolution.
Finally, we compare how human authors and machine translations handle CSIs, including the strategies they employ, by manually assigning strategy label(s) to each culture-specific AD.

\mypara{Results and discussion.}
\begin{table}[t]
\centering
\tabcolsep=0.16cm
\small
\caption{\textbf{CSI resolution rates across  categories.}
$\uparrow$ is better.
Human-authored Hindi ADs (\HIHuman) consistently outperform machine translated human-authored English ADs (\HumanADToHindi), with the highest resolution in cases of \colorbox{sameevent}{same detail} (N=63).
Lower resolution rates in \colorbox{partialevent}{partial} (N=39) and \colorbox{differentevent}{different} (N=18) cases suggests that humans adapt or avoid CSI-heavy content when direct resolution is difficult.}
\vspace{-1mm}
\begin{tabular}{lcccc}
\toprule
\textbf{Method} & \textbf{Aligned} & \textbf{\colorbox{sameevent}{Same}} & \textbf{\colorbox{partialevent}{Partial}} & \textbf{\colorbox{differentevent}{Different}} \\
\midrule
\HIHuman & 42.5\% & 50.8\% & 43.6\% & 11.1\% \\
\HumanADToHindi               & 10.0\% & 7.9\%  & 12.8\% & 11.1\% \\
\bottomrule
\end{tabular}
\label{tab:csi_results}
\vspace{-3mm}
\end{table}

We present results for CSI resolution in \cref{tab:csi_results}.
Overall, among the aligned ADs, we observe that \HIHuman ADs (42.5\% resolution rate) substantially outperform machine translation (10.0\% resolution rate) in resolving CSIs.
We study the resolution rate across different levels of overlapping detail and find that when humans preserve the original content (same details), they resolve CSIs at the highest rate (50.8\%), indicating deliberate effort to adapt cultural references.
In contrast, resolution drops in partial (43.6\%) and different (11.1\%) cases.
This suggests that when CSIs are difficult to adapt, humans may avoid resolving them and instead reframe the AD itself.
Machine translation, on the other hand, shows low resolution rates consistently (hovering around 10\%) across all categories.
This likely stems from the fact that the translations are not \textit{Skopos}-oriented, as they are designed to prioritize literal fidelity over adapting content to the target culture.

\begin{figure}[t]
\centering
\includegraphics[width=0.75\linewidth]{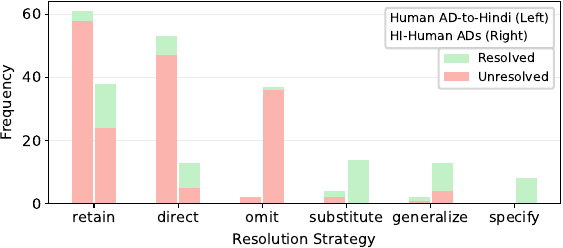}
\vspace{-3mm}
\caption{\textbf{CSI resolution strategies in \HumanADToHindi ADs (left) vs. \HIHuman ADs (right).}
Bars indicate the strategy used for resolved (green) and unresolved (red) CSIs.
Model translations (left) primarily rely on retention and direct translation and feature low resolution rates, while humans (right) use more adaptive strategies (\eg~substitute, generalize, specify) and resolve CSIs more effectively.
}
\label{fig:csi_plot}
\vspace{-5mm}
\end{figure}

We discuss the CSI adaptation strategies used by humans and machine translations in  \cref{app:csi_discussion}, with examples in \cref{fig:CSI_examples}.
We see that translated ADs (\HumanADToHindi) overwhelmingly rely on retention and direct translation, leaving most CSIs unresolved.
This conflicts with the purpose of the translation, Indian BLV accessibility.
In contrast, AD experts (\HIHuman) also employ retention and direct translation, but they resolve CSIs more frequently, indicating that these strategies can be used more effectively.
Translations, by design, almost never use omission. 
Meanwhile, humans rely heavily on it, even though it often results in unresolved CSIs.
This suggests that human AD experts omit details to avoid difficult-to-resolve CSIs or because such CSIs cannot be meaningfully adapted.
This is consistent with Toury~\cite{toury1995descriptive}, which identifies omission as a legitimate and frequent strategy.
For example, in subtitling, omission is sometimes the only feasible strategy~\cite{pedersen2011_csi_taxonomy}.
In general, we find that humans favor adaptive strategies such as substitution, generalization, and specification. 
These strategies show higher resolution rates because they reinterpret content for the target culture, aligning well with Skopos theory.
Examples of CSIs and their resolution strategies are presented in \cref{app:csi_discussion}.

In summary, translations are faithful to the source but unaware of purpose, whereas humans are more purpose-driven and culturally adaptive.
These findings suggest that translation systems, if used for creating ADs in Hindi, should adapt culturally grounded ADs to the needs of the target audience.

\subsection{Automatic translation of English human-authored ADs}
\label{subsec:trans_enhuman}
\begin{table}[t]
\centering
\small
\tabcolsep=0.15cm
\caption{\textbf{LLM-AD-Eval scores for human-authored ADs.}
We evaluate AD quality with two models as LLM-as-a-judge setup: Gemini-3.1-Pro~\cite{geminiteam2025geminifamilyhighlycapable} and LLaMA-3-8B-Instruct~\cite{grattafiori2024llama3herdmodels}.
We show results for two sets of ADs based on overlapping details:
\textit{All events} and \textit{Same+Partial events}. \textbf{Best results} and \underline{second-best} results are highlighted.
}
\label{tab:human_ad_quality}
\vspace{-1mm}
\begin{tabular}{llccccc}
\toprule
Detail & Judge &
IndicTrans2 & NLLB & MADLAD & Gemini & Gemini w/ ctxt \\
\midrule
\multirow{2}{*}{All}
  & Llama      & 3.98 & 3.91 & 3.61 & \underline{4.04} & \textbf{4.06} \\
  & Gemini  & 3.82 & 3.12 & 3.04 & \underline{4.11} & \textbf{4.14} \\
\midrule
\multirow{2}{*}{\makecell[l]{Same + \\ Partial}}
  & Llama       & 4.03 & 3.96 & 3.71 & \underline{4.08} & \textbf{4.12} \\
  & Gemini  & 3.99 & 3.82 & 3.21 & \underline{4.24} & \textbf{4.43} \\
\bottomrule
\end{tabular}
\vspace{-5mm}
\end{table}

While translating automatically generated ADs produced poor results (\cref{tab:machine_ad_quality_merged}), does automatic translation of \ENHuman ADs fare better?
We present results on 3 movies; \textit{Extraction 2} is excluded due to differences in Hindi/English ADs.

\mypara{LLM-AD-Eval.}
From \cref{tab:human_ad_quality}, we observe that machine-translated Hindi ADs do preserve the broad content of human-authored English ADs, resulting in scores close to 4 (5 being the highest).
However, they still do not match the quality of human-authored Hindi descriptions.
Gemini-3.1-Pro with previous AD/Dialog context outperforms all machine translation models. 
Scores are slightly higher for the \textit{Same+Partial details} setting because it removes cases where the two AD tracks (English and Hindi) focus on different salient details.
This also indicates that part of the challenge in the \textit{All details} setting arises from humans choosing to describe different details rather than translation quality.
Despite these high scores, recall that \cref{tab:csi_results} shows that translations resolve only 10\% of CSIs compared to 42.5\% by human-authored ADs.
Thus, while LLM-AD-Eval captures surface-level semantic overlap, it ignores whether the translation fulfills its purpose for the target audience.

\begin{figure}[!b]
\centering
\vspace{-4mm}
\includegraphics[width=0.81\linewidth]{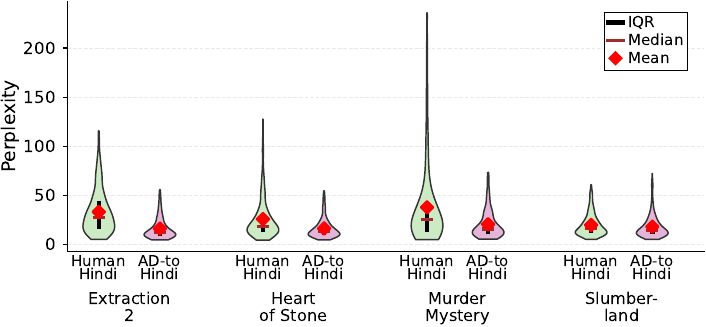}
\vspace{-2mm}
\caption{\textbf{Perplexity of human-authored Hindi ADs and machine translations of human-authored English ADs.} 
Perplexity of translated English ADs is lower than human-authored Hindi ADs across all movies with dual AD tracks.
IndicTrans2 is used to perform translation.
The base model is Llama-3.1-8B.
The plot shows values up to the 95th percentile.}
\label{fig:human_authored_perplexity}
\end{figure}

\mypara{Perplexity.}
We report diversity of the ADs of both tracks (\HIHuman and \HumanADToHindi) in~\cref{fig:human_authored_perplexity}.
Similar to \cref{sec:generation}, we observe that machine translating human-authored English ADs to Hindi results in reduced perplexity.
This suggests that translated ADs are more predictable, and are likely less diverse compared to human-authored Hindi ADs.

\section{Conclusion}
\label{sec: discussion_and_conclusion}
We presented the first systematic study of Hindi ADs, addressing the gap in AD research in the Indian context. 
Through the Andha-Dhun dataset and evaluation of translation-based approaches, we found that current methods remain significantly below human-authored quality, especially in handling cultural references and preserving linguistic diversity.
We find that directly generating Hindi ADs from dense descriptions can be more effective than translating predicted English ADs, indicating a promising direction for future research.
Our findings also identified that the true purpose of Hindi ADs, to make Indian BLV audiences understand the movie and its visuals, requires prioritizing adaptation over strict translation.
This hints that we need to develop methods that incorporate cultural and audience-aware adaptation.

\mypara{Acknowledgments.}
This project was supported by SERB SRG/2023/002544, Adobe Research India, and Amazon Prime Video.
We thank Google for sponsoring Gemini API credits.

\bibliographystyle{splncs04}
\bibliography{bib/shortstrings,bib/refs}

\clearpage
\appendix

\section{Supplementary Material}
\label{app:supplementary}

We describe in detail the prompts for each generation/evaluation pipeline.

\subsection{Dataset}
\label{app:dataset}

\mypara{Prompt for \dataset{} Transcriptions.}
We use Gemini-2.5-Flash to transcribe Hindi AD audio files into timestamped, speaker-classified annotations. The prompt instructs the model to distinguish between narrator speech (audio descriptions) and actor speech (dialogs), producing structured 4-column output. The full prompt is shown in \cref{fig:gemini_transcription_prompt}.

\begin{figure*}[h]
\centering
\normalsize
\noindent\begin{minipage}{1.0\linewidth}
\mdfsetup{middlelinewidth=1pt,backgroundcolor=green!3,innerleftmargin=0.4cm,innerrightmargin=0.4cm,roundcorner=10pt}
\begin{mdframed}
\setlength{\parskip}{6pt}
You are given a Hindi audio file that contains a mixture of narration and dialogues.

Your task is to generate structured annotations in the following 4-column format: [start\_time] [end\_time] [Speaker: Narrator or Actor] [Text]

Instructions:
\begin{enumerate}[noitemsep,topsep=2pt,parsep=0pt,partopsep=0pt]
    \item Classify each line of speech as either spoken by a ``Narrator'' or an ``Actor''.
    \item Maintain continuous timestamps from start to finish (timestamps should never reset).
    \item Text must be the spoken line in Hindi, using Devanagari script.
    \item Do not include speaker names such as ``Farhaan'', ``Raaju'', etc. Just use ``Narrator'' or ``Actor''.
    \item Ensure each annotation has accurate start and end times in the format minutes+seconds+milliseconds (e.g., 02:00.010 to 02:00.050).
    \item The output should follow the style shown below:\footnotemark
\end{enumerate}

\vspace{2pt}
\texttt{01:59.98~~02:02.66~~Narrator:~~Usse jhempte hue apni jeb se phone nikala, number dekha.}\\
\texttt{02:03.11~~02:04.14~~Actor:~~Hello?}
\vspace{2pt}

Do not include any extra commentary, metadata, character names, or formatting other than what is asked above.
\end{mdframed}
\end{minipage}
\footnotetext{Hindi examples shown in romanized form; the actual prompt uses Devanagari script.}
\caption{Prompt used with Gemini-2.5-Flash for transcribing Hindi AD audio into timestamped, speaker-classified annotations. Hindi text is romanized for simplicity.}
\label{fig:gemini_transcription_prompt}
\end{figure*}

\clearpage
\subsection{Machine-Generated Audio Description}
\label{app:machine_generated_ad}

\mypara{Dense-to-Hindi Generation Prompt.}
For \textit{Dense-to-Hindi} AD generation, we use the following system prompt with both Nemotron-Mini-Hindi-4B-Instruct and Gemini-3.1-Pro. The prompt instructs the model to convert English dense visual descriptions into natural, narrator-style Hindi ADs in Devanagari script. The full prompt is shown in \cref{fig:dense_to_hindi_prompt}.

\begin{figure*}[h]
\centering
\normalsize
\noindent\begin{minipage}{1.0\linewidth}
\mdfsetup{middlelinewidth=1pt,backgroundcolor=green!3,innerleftmargin=0.4cm,innerrightmargin=0.4cm,roundcorner=10pt}
\begin{mdframed}
\setlength{\parskip}{6pt}
You are a professional writer for film audio description (AD) in Hindi.

Your task is to convert an English dense visual description of a movie clip into natural, fluent, narrator-style Hindi audio description in Devanagari script.

Important requirements:
\begin{enumerate}[noitemsep,topsep=2pt,parsep=0pt,partopsep=0pt]
    \item This is for FILM AUDIO DESCRIPTION, not subtitles or dialogue.
    \item Preserve the visible meaning from the input and do not hallucinate details.
    \item Keep the AD concise, natural, and suitable for spoken narration.
    \item Focus on the most attractive / salient characters and their actions.
    \item For characters, use their first names when available, and remove titles such as `Mr.' and `Dr.'.
    \item If names are not available, use natural pronouns where possible; avoid generic phrases like `a man' unless truly necessary.
    \item For actions, avoid mentioning the camera, and do not focus on talking or position-related actions such as sitting and standing unless essential.
    \item Do not mention characters' mood unless absolutely necessary from the visible description.
    \item Prefer natural spoken Hindi, not stiff or overly literal Hindi.
    \item If a term does not have a good Hindi equivalent in this film context, keep it as a natural English-origin word written in Devanagari.
    \item Adjust output length according to the clip duration.
    \item Return valid JSON only in exactly this format: \{``summarised\_AD'': ``<YOUR OUTPUT>''\}
\end{enumerate}
\end{mdframed}
\end{minipage}
\caption{System prompt used for \textit{Dense-to-Hindi} AD generation with Nemotron and Gemini.}
\label{fig:dense_to_hindi_prompt}
\end{figure*}

\mypara{LLM-AD-Eval Prompt.}
We use the LLM-AD-Eval framework to evaluate the semantic match between predicted and ground-truth ADs. The system prompt, shown in \cref{fig:llm_ad_eval_prompt}, instructs the judge model to focus on visual elements and assign a score from 0 (worst) to 5 (best).

\begin{figure*}[t]
\centering
\normalsize
\noindent\begin{minipage}{1.0\linewidth}
\mdfsetup{middlelinewidth=1pt,backgroundcolor=green!3,innerleftmargin=0.4cm,innerrightmargin=0.4cm,roundcorner=10pt}
\begin{mdframed}
\setlength{\parskip}{6pt}
You are an intelligent chatbot designed for evaluating the quality of generative outputs for movie audio descriptions. Your task is to compare the predicted audio descriptions with the correct audio descriptions and determine its level of match, considering mainly the visual elements like actions, objects and interactions. Here's how you can accomplish the task:

Instructions:
\begin{itemize}[noitemsep,topsep=2pt,parsep=0pt,partopsep=0pt]
    \item Check if the predicted audio description covers the main visual events from the movie, especially focusing on the verbs and nouns.
    \item Evaluate whether the predicted audio description includes specific details rather than just generic points. It should provide comprehensive information that is tied to specific elements of the video.
    \item Consider synonyms or paraphrases as valid matches. Consider pronouns like `he' or `she' as valid matches with character names. Consider different character names as valid matches.
    \item Provide a single evaluation score that reflects the level of match of the prediction, considering the visual elements like actions, objects and interactions.
\end{itemize}
\end{mdframed}
\end{minipage}
\caption{System prompt for LLM-AD-Eval, following the same evaluation setup as \cite{autoadzero}. The LLM judge assigns a score from 0 (worst) to 5 (best) reflecting the semantic match between a predicted AD and the ground-truth AD.}
\label{fig:llm_ad_eval_prompt}
\end{figure*}

\subsection{Human-Authored Audio Description Analysis}
\label{app:human_authored_ad_generation}

\mypara{Detail Overlap Prompt.}
To classify the detail overlap between \textit{HI-Human} and \textit{Human AD-to-Hindi} AD pairs, we use the prompt shown in \cref{fig:detail_overlap_prompt}.

\begin{figure*}[t]
\centering
\normalsize
\noindent\begin{minipage}{1.0\linewidth}
\mdfsetup{middlelinewidth=1pt,backgroundcolor=green!3,innerleftmargin=0.4cm,innerrightmargin=0.4cm,roundcorner=10pt}
\begin{mdframed}
\setlength{\parskip}{6pt}
You are an expert bilingual annotator for Hindi movie audio description (AD) tracks. Audio descriptions narrate on-screen visual events for visually impaired viewers.

You will receive two inputs per row:

\texttt{HI\_HUMAN}~~~: human-authored ground-truth Hindi AD\\
\texttt{HUMAN\_AD\_TO\_HINDI}: machine-translated Hindi AD from human-authored English AD

\textbf{TASK: detail\_overlap}

Decide whether the two Hindi sentences describe the SAME visual details.

Focus ONLY on visual content: actions, objects, characters, spatial relations. Ignore style, level of detail, and transliteration variants --- ``\textit{Taaylar}'' and ``\textit{Taailar}'' are the same character Tyler; ``\textit{Ejeebeeoh}'' and ``\textit{Egbo}'' are the same production house; ``\textit{vah}'' standing in for a named character is not a mismatch.

Labels:
\begin{itemize}[noitemsep,topsep=2pt,parsep=0pt,partopsep=0pt]
    \item \texttt{SAME\_DETAILS} --- same primary action/event even if one sentence is more detailed or uses synonyms/paraphrases.
    \item \texttt{PARTIAL\_OVERLAP} --- share some visual content but one sentence also describes a clearly distinct additional action absent from the other.
    \item \texttt{DIFFERENT\_DETAILS} --- completely unrelated details with no meaningful overlap in action, object, or character.
\end{itemize}

\textbf{OUTPUT FORMAT}

Return ONLY a JSON object with exactly these two keys --- no extra text:

\vspace{2pt}
\texttt{\{}\\
~~\texttt{"detail\_overlap":~~~"<SAME\_DETAILS | PARTIAL\_OVERLAP | DIFFERENT\_DETAILS>",}\\
~~\texttt{"overlap\_reason":~~"<one concise sentence>"}\\
\texttt{\}}
\end{mdframed}
\end{minipage}
\caption{Prompt used with Gemini-3.1-Pro for classifying detail overlap between \textit{HI-Human} and \textit{Human AD-to-Hindi} AD pairs.}
\label{fig:detail_overlap_prompt}
\end{figure*}

\begin{figure}[t]
\centering
\includegraphics[width=0.9\linewidth]{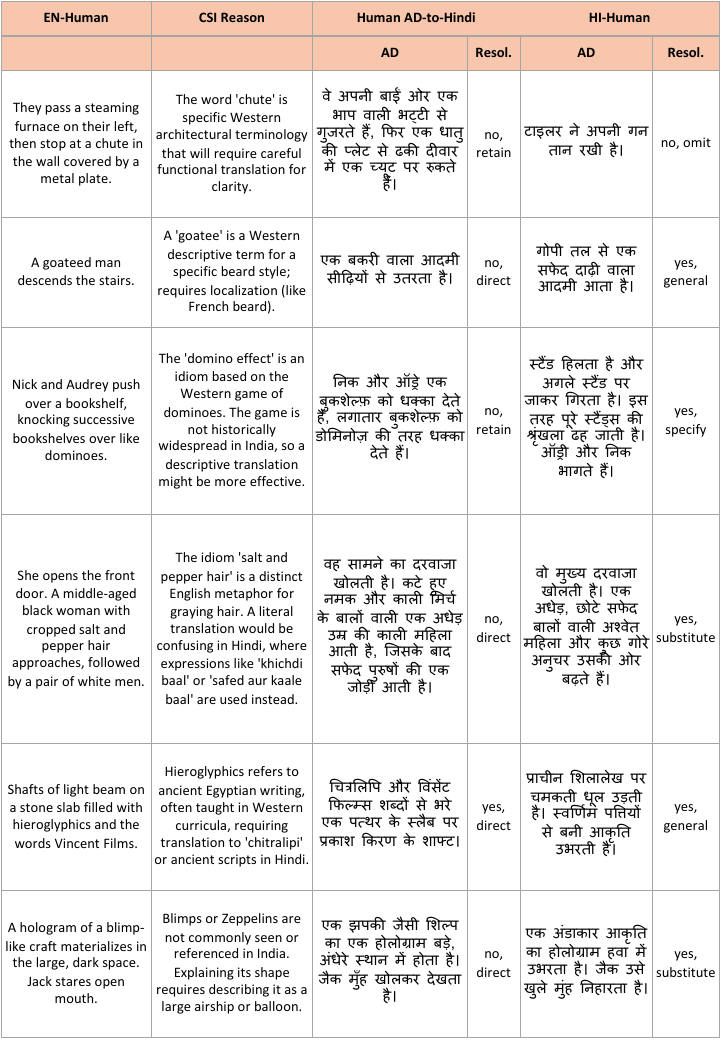}
\caption{\textbf{Examples of Culture-specific Items and their resolutions.} Each row presents the source English AD, the identified CSI with explanation, the machine-translated Hindi AD from English AD, and the human-authored Hindi AD, along with resolution outcomes and strategies. The figure shows how different strategies are applied in practice to either resolve CSIs or leave them unresolved.}
\label{fig:CSI_examples}
\end{figure}

\mypara{CSI Detection and Examples.}
\label{app:csi_discussion}
To identify Culture-Specific Items (CSIs) in English ADs, we use the prompt shown in \cref{fig:csi_detection_prompt}. Examples of identified CSIs and their resolution strategies across human-authored and machine-translated Hindi ADs are provided in \cref{fig:CSI_examples}.

\begin{figure*}[t]
\centering
\normalsize
\noindent\begin{minipage}{1.0\linewidth}
\mdfsetup{middlelinewidth=1pt,backgroundcolor=green!3,innerleftmargin=0.4cm,innerrightmargin=0.4cm,roundcorner=10pt}
\begin{mdframed}
\setlength{\parskip}{6pt}
You are an expert Audio Description (AD) translator and cultural consultant.  
We are translating English Audio Descriptions into Hindi for an Indian Blind and Low Vision (BLV) audience.  
We apply Skopos theory, meaning the translation's primary purpose is to be functionally effective and culturally accessible to the target audience, rather than a literal word-for-word translation.

Your task is to analyze a batch of English Audio Descriptions and determine if each contains any Culture-Specific Items (CSIs) that might not be easily understood by an Indian BLV audience.

Examples of CSIs between English and Hindi contexts:
\begin{itemize}[noitemsep,topsep=0pt,parsep=0pt,partopsep=0pt]
    \item ``He scored a touchdown in the final quarter.'' (American football reference)
    \item ``They celebrated Thanksgiving with turkey and pumpkin pie.'' (American holiday and food)
    \item ``He's applying to Ivy League colleges.'' (US education system)
    \item ``They watched the Queen's Speech on Christmas Day.'' (British cultural event)
    \item ``They had fish and chips by the seaside.'' (British food)
\end{itemize}

Here is the batch of English ADs to evaluate, provided as a JSON array of strings:
\{batch\_json\}

Evaluate each AD and return a JSON array of objects. Each object must have:
\begin{enumerate}[noitemsep,topsep=2pt,parsep=0pt,partopsep=0pt]
    \item ``ad\_text'': The exact English AD text from the input.
    \item ``is\_csi'': ``yes'' if the AD contains a cultural element likely unfamiliar to the target audience, or ``no'' if it is culturally neutral or globally understood.
    \item ``reason'': A brief explanation of your decision.
\end{enumerate}

Respond ONLY with a valid JSON array of objects.
\end{mdframed}
\end{minipage}
\caption{Prompt for detecting Culture-Specific Items (CSIs) in English to Hindi AD translation.}
\label{fig:csi_detection_prompt}
\end{figure*}

\end{document}